\documentclass[runningheads]{ks_press}
\usepackage[T1]{fontenc}
\usepackage{amsmath}
\usepackage{amsfonts}
\usepackage{amssymb}
\usepackage{hyperref}
\usepackage{graphicx}
\usepackage{romannum}
\usepackage{booktabs} % For pretty tables
\usepackage[all]{nowidow}
\usepackage[utf8]{inputenc}
\usepackage{gensymb}
\usepackage{float}

\usepackage[mathlines]{lineno}
\usepackage{etoolbox} %% <- for \pretocmd, \apptocmd and \patchcmd
\newcommand*\linenomathpatchAMS[1]{%
  \expandafter\pretocmd\csname #1\endcsname {\linenomathAMS}{}{}%
  \expandafter\pretocmd\csname #1*\endcsname{\linenomathAMS}{}{}%
  \expandafter\apptocmd\csname end#1\endcsname {\endlinenomath}{}{}%
  \expandafter\apptocmd\csname end#1*\endcsname{\endlinenomath}{}{}%
}

\expandafter\ifx\linenomath\linenomathWithnumbers
  \let\linenomathAMS\linenomathWithnumbers
  
  \patchcmd\linenomathAMS{\advance\postdisplaypenalty\linenopenalty}{}{}{}
\else
  \let\linenomathAMS\linenomathNonumbers
\fi

\linenomathpatchAMS{gather}
\linenomathpatchAMS{multline}
\linenomathpatchAMS{align}
\linenomathpatchAMS{alignat}
\linenomathpatchAMS{flalign}

\makeatletter
\renewcommand\section{\@startsection{section}{1}{\z@}%
                       {-8\p@ \@plus -4\p@ \@minus -4\p@}%
                       {6\p@ \@plus 4\p@ \@minus 4\p@}%
                       {\normalfont\large\bfseries\boldmath
                        \rightskip=\z@ \@plus 8em\pretolerance=10000 }}
\renewcommand\subsection{\@startsection{subsection}{2}{\z@}%
                       {-8\p@ \@plus -4\p@ \@minus -4\p@}%
                       {6\p@ \@plus 4\p@ \@minus 4\p@}%
                       {\normalfont\normalsize\bfseries\boldmath
                        \rightskip=\z@ \@plus 8em\pretolerance=10000 }}
\renewcommand\subsubsection{\@startsection{subsubsection}{3}{\z@}%
                       {-4\p@ \@plus -4\p@ \@minus -4\p@}%
                       {-1.5em \@plus -0.22em \@minus -0.1em}%
                       {\normalfont\normalsize\bfseries\boldmath}}
\makeatother
\usepackage[algoruled]{algorithm2e}

\begin{document}
%%%%%%%%%%%%%%%%%%%%%%%%%%%%%%%%%%%%%%%%%
\setpagewiselinenumbers
\setcounter{page}{47}
%\linenumbers %to be included for Proof

%%%%%%%%%%%%%%%%%%%%%%%%%%%%%%%%%%%%%%%%%%%%%%%

%%%%%%%%%%%%%%%%%%%%%%%%%%%%%%

\journal{International Journal of Robotic Computing}
\volume{Vol. 3, No. 1 (2021) 47-68}
\publisher{© KS Press, Institute for Semantic Computing Foundation}
\myDOI{10.35708/RC1870-126265}


\title{Reactive Navigation Framework for Mobile Robots by Heuristically Evaluated Pre-sampled Trajectories}
\titlerunning{Reactive Navigation Framework for Mobile Robots}

\author{Ne\c{s}et \"{U}nver Akmandor \inst{1} \and Ta\c{s}k{\i}n Pad{\i}r \inst{2}}
\authorrunning{N.U. Akmandor et al.}
\institute{Department of Electrical and Computer Engineering\\
Northeastern University, Boston, MA, 02115, USA\\
\email{akmandor.n@northeastern.edu}
\and Institute for Experiential Robotics\\
Northeastern University, Boston, MA, USA\\
\email{t.padir@northeastern.edu}
}
\maketitle

%%%%% History : to be added by Publisher %%%%%%%
\begin{history}
\received{(03/10/2021)}
%\revised{(xx/xx/xxxx)}
\accepted{(03/26/2021)}
\end{history}
%%%%%%%%%%%%%%%%%%%%%%%%%%%%%%%%%%%%%%%%%%%%%%%%%%%%%%%

\begin{abstract}

This paper describes and analyzes a reactive navigation framework for mobile robots in unknown environments. The approach does not rely on a global map and only considers the local occupancy in its robot-centered 3D grid structure. The proposed algorithm enables fast navigation by heuristic evaluations of pre-sampled trajectories on-the-fly. At each cycle, these paths are evaluated by a weighted cost function, based on heuristic features such as closeness to the goal, previously selected trajectories, and nearby obstacles. This paper introduces a systematic method to calculate a feasible pose on the selected trajectory, before sending it to the controller for the motion execution. Defining the structures in the framework and providing the implementation details, the paper also explains how to adjust its offline and online parameters. To demonstrate the versatility and adaptability of the algorithm in unknown environments, physics-based simulations on various maps are presented. Benchmark tests show the superior performance of the proposed algorithm over its previous iteration and another state-of-art method. The open-source implementation of the algorithm and the benchmark data can be found at \url{https://github.com/RIVeR-Lab/tentabot}.
\end{abstract}

\keywords{reactive navigation; trajectory sampling; heuristic functions; robot-centered 3D grid; local mapping}

%-------------------------------------------------------------------------
\section{Introduction}
\label{S:intro}
Despite being studied for more than few decades, motion and path planning literature still embodies lots of challenging and open-ended research topics such as mapping and localization \cite{oleynikova2019open,droeschel2017continuous}, exploration \cite{oleynikova2018safe,faria2019efficient} and safe path planning \cite{tordesillas2020faster,gao2018online,lin2018autonomous,usenko2017real,oleynikova2016continuous}. Considering their mobility and flexibility, the real-world applications with unmanned aerial vehicles (UAV) and autonomous underwater vehicles (AUV) have become the focus of many academic \cite{lin2018autonomous,oleynikova2016continuous,pereira2013risk} or industrial projects \cite{beul2018fast,escobar2018r}.
Based on their use cases, these applications commonly require considerable amount of memory to store the data and computational power to process them. In order to meet these requirements, researchers \cite{campos2019autonomous,mohta2018fast} need to develop systems and algorithms that are capable of working onboard and processing online data. ~\\

\subsection{Contribution}

Assuming a fully observable global map is highly impractical for a real-world autonomous navigation application. To navigate in unknown, highly cluttered and dynamical environments, the algorithm needs to adapt fast enough to respond the variations around the autonomous agent. In this context, we propose a reactive navigation framework which does not use any prior global map information. Throughout the navigation, the local map around the robot is kept and updated by the incoming sensor data. The robot-centered 3-dimensional grid structure enables fast queries to extract the occupancy information around the pre-sampled trajectories. At each cycle, the algorithm computes the robot's next pose by evaluating the heuristic functions for each trajectory. ~\\
\\
In this paper, the sections \ref{sec:context}, \ref{sec:3dgrid}, \ref{sec:sample_traj}, \ref{sec:support_priority}, \ref{sec:traj_eval}, \ref{sec:traj_sel} and \ref{sec:parameters} explain the concepts introduced in \cite{akmandor20203d} in more details. Sec. \ref{sec:local_map} introduces the local map feature added into the proposed framework. We provide the updates on two of the heuristic functions, which are normalized, within the section \ref{sec:traj_eval}. A new systematic approach in Sec. \ref{sec:next_pose} is added with its algorithm to calculate feasible robot pose before it is sent to the controller for the motion execution. Considering the new features, such as local map and next pose calculation, the main algorithm and its implementation details are updated and discussed further in \ref{sec:implement}. In Sec. \ref{sec:results}, benchmark simulations are performed for the updated algorithm and their results are added into the previous results' plots. Based on the performance metrics, our proposed algorithm outperforms the state-of-art method as well as our previous implementation. To enable future benchmarks with our algorithm, the open-source implementation of the recent version and datasets are provided at \url{https://github.com/RIVeR-Lab/tentabot}. ~\\

\subsection{Related Work}
Considering approaches that enable autonomous navigation in an unknown 3D space, the state-of-art methods can be mainly categorized into four groups: optimization-based, sampling-based, reactive and the compounds which are obtained by the algorithmic fusion of the aforementioned categories. Each group has its own  advantages/disadvantages and their success/failure highly depends on the given task. In general, optimization-based methods suffer from the local minima while their mathematical groundings provide the robustness. Sampling-based algorithms are capable of finding global solutions. However, reaching that solution may take considerable amount of time depending on the dimensionality of the sampling space. Without having a planning step, reactive methods act upon their pre-defined behaviour given the sensor data. Hence, they are computationally fast, but their success highly depend on the designed policy and its interpretation of the sensor data. In between the foregoing groups, the compound algorithms is designed hierarchically to maximize the efficiency in a given task. However, this crafted design complicates the implementation of the algorithm and affects its generalizability to diverse applications. ~\\
\\
One of the related work within our problem definition \cite{usenko2017real} stores the local occupancy information around the robot in a 3D circular buffer and repeatedly adjust the local trajectory represented by a B-spline. Despite having the possibility of getting stuck at the local minima, the parameters of the B-spline is calculated by optimizing a cost function which pulls the robot towards goal and drives away from obstacles while keeping the robot's motion stable. The algorithms in Lin et al. \cite{lin2018autonomous} and Gao et al. \cite{gao2018online} require high computation power due to their image processing and optimization steps. Both frameworks estimate the 3D local map using the camera and inertial measurement unit. Using the map, the work in \cite{lin2018autonomous} generates a local path by a sampling-based algorithm, RRG \cite{karaman2011sampling}, while \cite{gao2018online} uses fast marching method to obtain it. Initializing with the computed path, the non-linear optimization solver ensures the smoothness and dynamical feasibility of the final trajectory for each method. In \cite{mohta2018fast}, Mohta et al. propose a trajectory planner in GPS-denied and cluttered environments, providing detailed aspects on both hardware and software. Similar to the aforementioned algorithms, they also combine a sampling-based method, A* \cite{hart1968formal}, with an optimization process to generate the robot trajectory. To avoid local minima during the trajectory calculation, they propose a combined map structure which keeps the local occupancy information in 3D while the global one is in 2D. However, even though the global map is planar, the size of the map and discrete nature of the A* algorithm limit the applicability of their framework to real-world scenarios. ~\\
\\
In their work \cite{oleynikova2019open}, Oleynikova et al. propose a framework for mapping, planning and trajectory generation. Having a vision based sensing, the Truncated Signed Distance Field (TSDF) is computed to project the environment around the robot into a map which represents collision costs. Generating a deterministic graph in the free-space of the map, the path is generated using the A* algorithm. In the last step, the trajectory of the robot is calculated by an optimization considering the trade-off between reaching to the goal and exploration. As a motion planning framework for the Micro Aerial Vehicles (MAVs), Campos-Macías et al. \cite{campos2019autonomous} represent the local occupancy information using a linear octree structure. Then, the path of the robot is planned by RRT-Connect \cite{karaman2011sampling}. The trajectory generation, which includes an offline stage of LQR virtual control design and Lyapunov analysis, guarantees that the dynamic constraints are satisfied. Kinodynamically sampling the space using machine learning and smoothing their resultant path, Allen et al. \cite{allen2019real} achieves real-time motion planning avoiding dynamical obstacles. Tordesillas et al.  \cite{tordesillas2020faster} use an optimization-based method to calculate two trajectories in each time step, by prioritizing safety for one of them and speed for the other. Then, these trajectories compromise to obtain a committed trajectory. Authors show that their approach can achieve high speed navigation in unknown environments. In another recent work \cite{zhou2020robust}, Zhou et al. also propose an optimization based approach in which re-planning is used to eliminate the local minima issue. ~\\
\\
As one of the earlier works in reactive control, Ulrich and Borenstein, \cite{ulrich2000vfh} present a local obstacle avoidance algorithm. Similar to our approach, they sample some feasible trajectories and evaluated them by a cost function. Then, the candidate direction is selected to guide the robot around obstacles. In \cite{von2008driving}, the 2007 European Land Robot Trial winner and DARPA Urban Challenge finalist team propose a reactive navigation algorithm, for their car-like robot, which enables fast navigation towards to a goal while avoiding obstacles in highly cluttered environments. In their paper, Von Hundelshausen et al. refer pre-calculated trajectories as tentacles which are formed in robot's coordinate frame. Their algorithm also considers these tentacles as perceptual primitives by mapping the occupancy grid information onto them. Later, they extend their work by accumulating LIDAR data into a multi-layered occupancy grid in \cite{himmelsbach2009team} and updating their circular tentacle form to clothoid considering steering angle in \cite{himmelsbach2011autonomous}. Integrating robot kinematics into circular tentacle calculation, Cherubini et al. \cite{cherubini2012new} use visual data to avoid static obstacles during the navigation. In their following paper \cite{cherubini2014autonomous}, the algorithm is extended by an obstacle observer model to enable dynamic obstacle avoidance. The work in \cite{alia2015local} forms clothoid version of tentacles and the selected tentacle is executed by their vehicle using a lateral controller based on ``Immersion and Invariance'' principle. Similarly by forming clothoid trajectories, the method in \cite{mouhagir2016markov} decides the best tentacle at each step by the Markov Decision Process. Following that work in \cite{mouhagir2017trajectory}, they introduce an ``evidential'' occupancy grid structure to represent sensor based uncertainties. In \cite{zhang2017formation}, Zhang et al. use tentacle concept to achieve multiple UAV formation flight and reactively reactive avoid obstacles. Most recently, Khelloufi et al. \cite{khelloufi2017tentacle} propose a tentacle-based obstacle avoidance scheme, for omni-directional mobile robots, which can visually track a target while navigating. ~\\
\\
Being reactive but not tentacle-based, the algorithms in \cite{blanco2008extending} and \cite{jaimez2015efficient} map the robot's workspace to a lower dimensional representations which is named as trajectory-parameter (TP). The pre-defined funtions are used to evaluate the TP-space image. Based on the evaluation, the corresponding robot action is selected to perform obstacle avoidance. Escobar et al. \cite{escobar2018r} and Beul et al. \cite{beul2018fast} use visual perception and reactive control algorithms to avoid obstacles and achieve fast navigation towards the goal with UAVs. Distinctively, Escobar et al. use potential fields to reach the goal, while Beul et al. plan a path of poses using the integration of A* and Ramer-Douglas Peucker algorithms. 

\section{3D Reactive Navigation Framework}

\subsection{Context} \label{sec:context} 

Our navigation framework is defined in 3D workspace which consists of either free or occupied subspaces in a fixed Cartesian coordinate frame $W$. The occupied subspace contains both static and dynamic objects, including our robot in frame $R$. The occupancy information around the robot can be obtained by one or more sensors. Since the fusion of multiple sensor data is beyond our paper, we assume the data is coming from a single source whose coordinate frame is defined as $S$. ~\\
\\
The main objective of our algorithm is to find a navigable path from a start position and orientation $(p_{start}, q_{start})$ to a goal $(p_{goal}, q_{goal})$, given multiple objectives such as; closest proximity to the goal, collision-free path and minimum navigation time. The structure of positions and orientations are defined as $p={[x,y,z]} \in \mathbb{R}^3$, $q={[x,y,z,w]}$ as a quaternion, respectively and we address them together as pose throughout the paper.

\subsection{Local Map} \label{sec:local_map}
Being a sensory input in our framework, the point cloud data, $D$, is  received at some frequency $f^S$. This data could be obtained by any sensor or multiple sensor fusion which measure spatial occupancy information around the robot. The framework assumes to receive the point cloud data, including the coordinate $^Sp_m$ of an occupied point $m$ with respect to sensor frame $S$, and $\rho_m$ as the probabilistic belief value of that point. The received point cloud data is transformed into the world coordinate frame and added into the local map around the robot. Considering that the minimum range of the sensor starts from some threshold, the local map plays a crucial role to avoid obstacles, especially when the robot changes its orientation rapidly. Having a local map is also more safer than keeping some history of the point cloud, since the data in the local map is updated based on spatial information but not the received order as in \cite{akmandor20203d}. ~\\
\\
In the literature, Octomap \cite{hornung2013octomap} and UfoMap \cite{duberg2020ufomap} are two open-source approaches that can be used to implement a map from the point cloud data. In \cite{duberg2020ufomap}, it is shown that UfoMap is more advantageous to represent the unknown region, more memory efficient and achieves faster insertion times than Octomap. However, we implement our local map as the Octomap structure for two reasons. First, in this paper we naively consider unknown regions as free of obstacles and leave to address this issue for a future work. Secondly, Octomap has better documentation at present. ~\\

\subsection{Robot-Centered 3D Grid} \label{sec:3dgrid}
\begin{figure}[!t]
\centering
\includegraphics[width=4.5in]{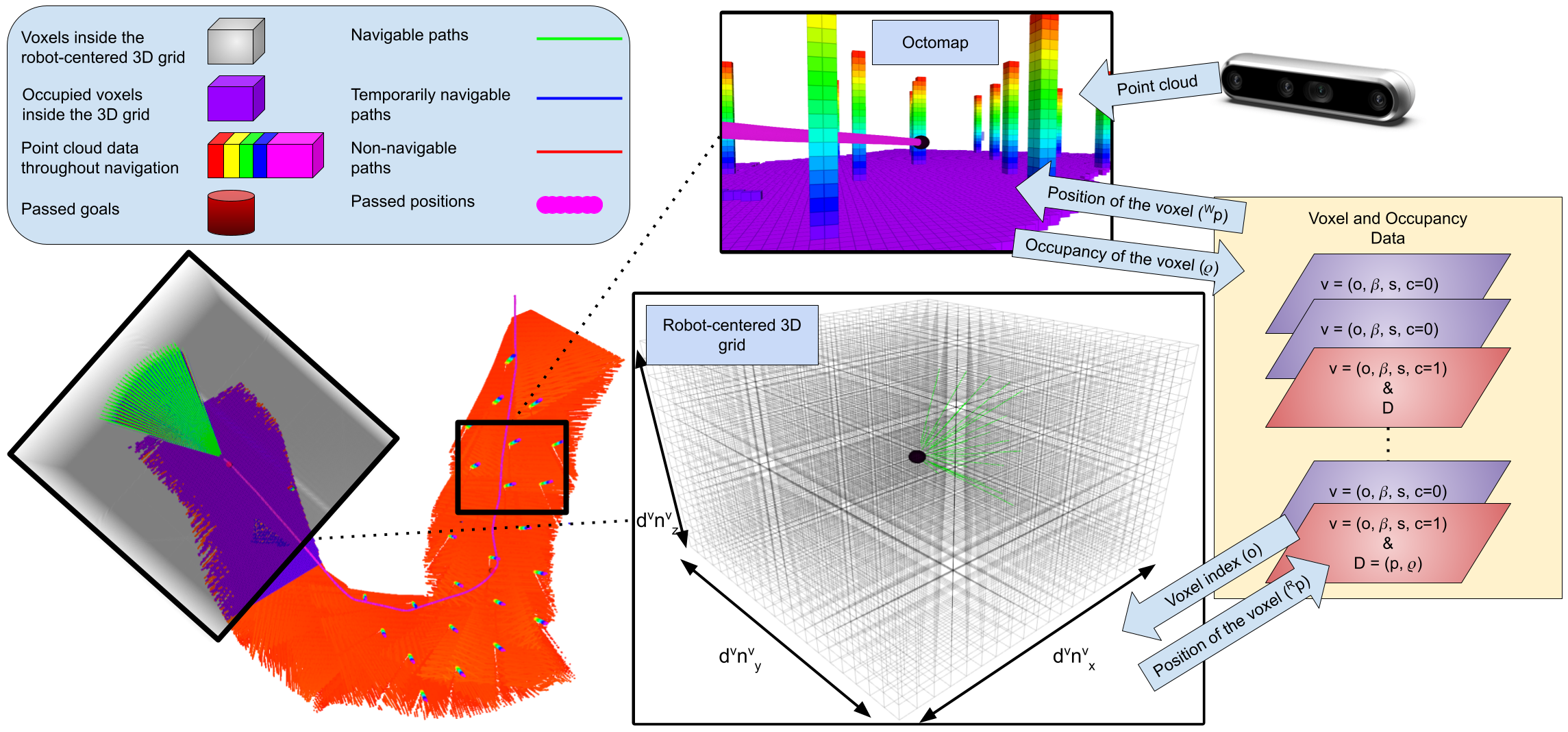}
\caption{The local map is updated by the point cloud data coming from the occupancy sensor. While navigating towards the goal, only obstacles inside the robot-centered grid (shaded grey region) is considered. The grid is formed in each axis of the robot's frame by $n^{v}_{\{x,y,z\}}$ number voxels with dimension $d^v$.}
\label{fig:local_map_robot_grid}
\end{figure}

Our previous paper \cite{akmandor20203d} extends the 2D approach in \cite{von2008driving} for the 3D case by forming a volumetric grid structure $G$, around the robot aligning with the robot's frame as shown in Fig. \ref{fig:local_map_robot_grid}. The robot-centered grid is composed of $N^{v}$ cubic voxels, determined by $N^{v} = n^{v}_z n^{v}_y n^{v}_z$, where $n^{v}_{\{x,y,z\}}$ are the number of voxels in each axes. The respective width, length and height ${\{w,l,h\}}^G$ of the grid is calculated as ${\{w,l,h\}}^G = d^{v} n^{v}_{\{x,y,z\}}$, given by the voxel dimension, $d^{v}$. ~\\
 ~\\
To enable fast query of occupancy info, the center position of each voxel in the 3D grid is mapped into a linear index. The mathematical formulation of this mapping, $M: \mathbb{R}^3 \rightarrow \mathbb{R}$ is given in the Eq. (\ref{eqn:lin}). 
\begin{subequations}
\label{eqn:lin}
\begin{align}
o_i &= o_{i_x} + o_{i_y} n^{v}_x + o_{i_z} n^{v}_x n^{v}_y \\
o_{i_{\{x,y,z\}}} &= \frac{n^{v}_{\{x,y,z\}}}{2} + floor(\frac{{\{x,y,z\}}}{d^v}). 
\end{align}
\end{subequations}
Having this mapping from Cartesian coordinates to the linear index, the array, $A_p$, stores the 3D positions with respect to the robot's frame $R$, while another array, $A_{\rho}$, keeps the occupancy info of all voxels. In order to update the occupancy info of a particular voxel $i$, the local map is queried by the transformation of $A_p(o_i)$ into the world frame $W$. ~\\
\begin{subequations}
\label{eqn:lin_arrays}
\begin{align}
A_p(o_i) &= ^{R}p_i\text{,} \\
A_{\rho}(o_i) &= \rho
\end{align}
\end{subequations}

\subsection{Sampling Trajectories and Navigation Points} \label{sec:sample_traj}
\begin{figure}[!t]
\centering
\includegraphics[width=4.5in]{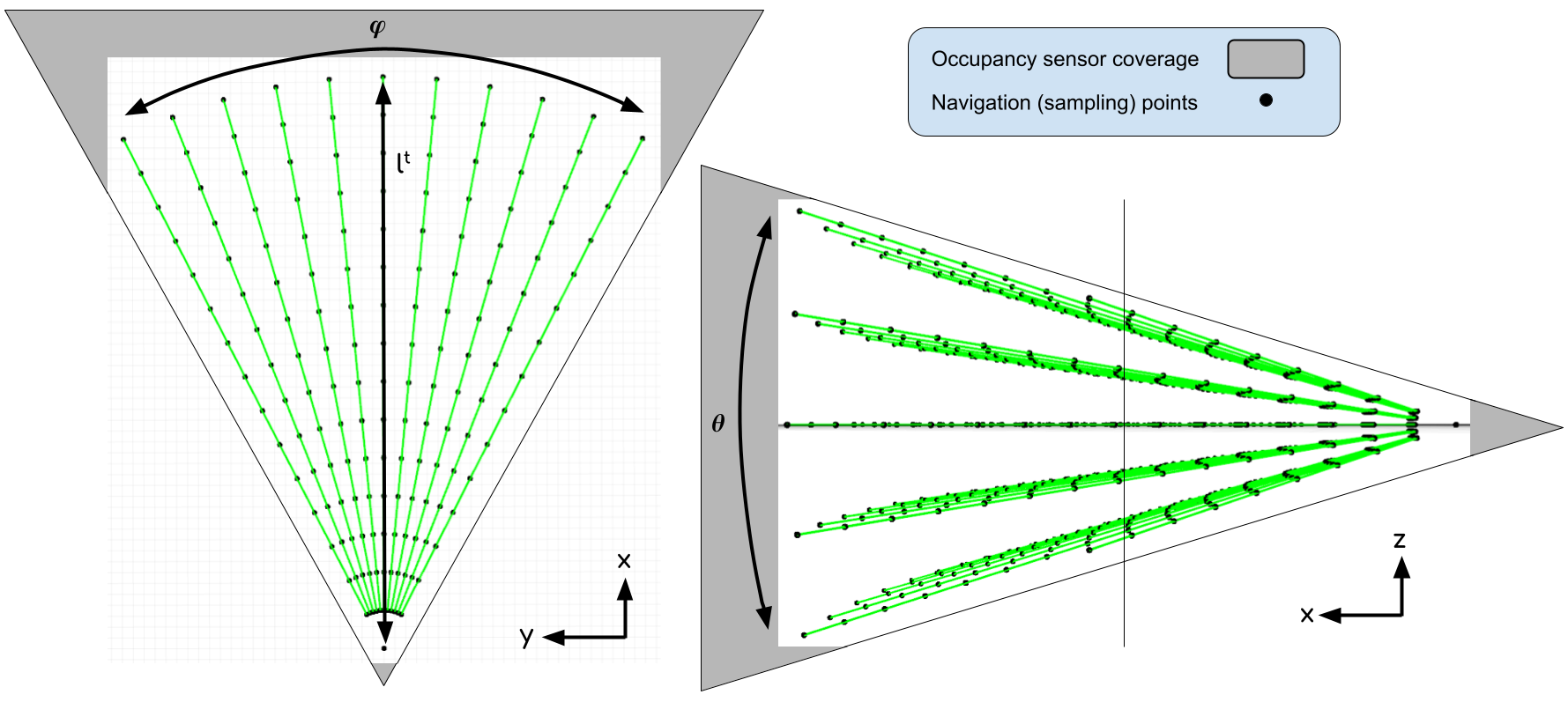}
\caption{Each trajectory is formed by the group of pre-calculated navigation points that are fixed to robot's coordinate frame. The length, $l^t_j$, of each trajectory $j$, yaw angle $\varphi$, and the pitch angle, $\theta$ are determined by the range of the occupancy sensor.}
\label{fig:tentacles}
\end{figure}

Simplifying their ground vehicle as the ``bicycle model'' and assuming constant lateral and angular velocities, \cite{von2008driving} generates their pre-calculated trajectories as circular arcs. When holonomic robots, such as UAVs, are considered, linear trajectories would be the logical choice since they sample the navigation space more uniformly than its counterparts and their geometric calculations are straightforward. ~\\
\\
Linear trajectories can be generated by calculating robot's motion for some time horizon while keeping the robot's orientation fixed on a particular sample from the possible orientation space. In our implementation, each trajectory has its own yaw and pitch angles which are obtained by sampling the angular coverage along the yaw $\varphi$ ($z$-axis), and the pitch $\theta$ ($y$-axis) as shown in Fig. \ref{fig:tentacles}. Hence, the total number of trajectories $N^t = n_{\varphi} n_{\theta}$ can be calculated by the multiplication of the number of samples along yaw $n_{\varphi}$, and pitch $n_{\theta}$. ~\\
\\
~\\
The length, $l^t_j$, of each trajectory $j$ is determined by the sensor range along its direction. Each trajectory is formed by $n^{s}$ number of navigation points in robot's coordinate frame. Similar to trajectory sampling, we linearly sample these points along their trajectory by having a constant $\delta d$ distance between each successive ones. These navigation points are linearly added until the sensor range is reached on the trajectory's direction. For each trajectory $j$, the position of each sampling point $^{R}p_{k_j}$ are stored in the set $T_j$. The total set, $Q$, of $N^t$ tentacles contains $N^s$ sampling points as shown in Eq. (\ref{eqn:tentacle_set}), where $N^s = N^t n^s_j$ and $n^s_j = l^t_j / \delta d$.
\begin{subequations}
    \begin{align}
        Q &= \{T_j \quad | \quad j = 1,...,N^t\} \\
        \quad T_j &= \{^{R}p_{k_j} \quad | \quad k_j = 1,...,n^s_j\}.
    \end{align}
\label{eqn:tentacle_set}
\end{subequations} ~\\
Generating these trajectories and sampling the navigation points by considering the dynamical structure of the robotic platforms tends to improve the performance of our navigation algorithm. However, that is not strictly necessary since the navigation points are also used to sense the environment and they are not directly sent to the motion execution. Therefore, instead of having kinodynamically sampled trajectories, the feasible navigation point on the selected trajectory is calculated before the motion execution in our framework. ~\\

\subsection{Support and Priority Voxels} \label{sec:support_priority}
\begin{figure}[!ht]
\centering
\includegraphics[width=4.5in]{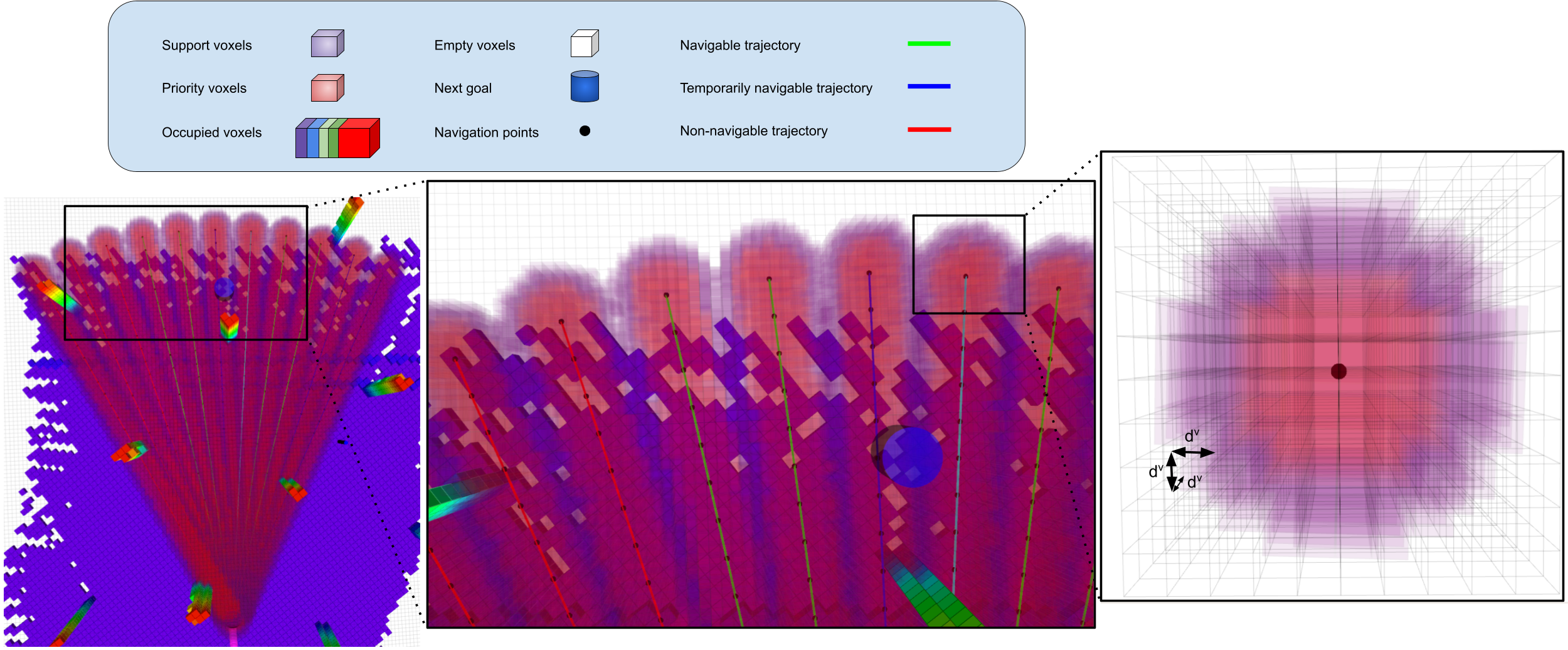}
\caption{Each trajectory has its own set of Support $S^v$ (magenta) and Priority $P^v$ (red) voxels inside the robot-centered grid. Trajectories are evaluated based on the occupancy in these voxels. If the occupied voxel is in $S^v$, its weight $\beta$ has higher value when it is closer to the trajectory. The weight gets its maximum when the voxel is in $P^v$. The occupancy around the robot determines whether the trajectory is navigable (green), non-navigable (red) or temporarily navigable (blue).}
\label{fig:support_priority}
\end{figure}

In the implementation of our framework, the voxel structure $v=(o, \beta, m, c)$ consists of four variables which are adjusted offline to enable fast computation of the heuristic values. The first variable $o$ is the index that points out the corresponding voxel position in the array $A_p$. The second variable $\beta$ keeps the occupancy weight based on the shortest distance between the voxel and the $j^{th}$ trajectory. The third variable $m$ holds the index of the closest navigation point on the $j^{th}$ trajectory to the voxel. Last variable $c$ indicates the class type of the voxel which can be either Priority ($c = 1$) or Support ($c = 0$). ~\\
\\
For each trajectory $j$, the subset of voxels inside the 3D grid are classified as either Support $S^{v}$ or Priority $P^{v}$ based on their closeness to the corresponding trajectory. These voxels are extracted as in the Eq. (\ref{eqn:sp_voxel}), given the distance thresholds ${\tau^{S^{v}}}$ and ${\tau^{P^{v}}}$ which needs to be ${\tau^{S^{v}}} > {\tau^{P^{v}}}$. Basically, if the distance between a particular voxel $i$ and the closest navigation point $p_{m_i}$ is less than ${\tau^{P^{v}}}$, that voxel is considered as Priority. If that distance is between ${\tau^{S^{v}}}$ and ${\tau^{P^{v}}}$, then the voxel is classified as Support. Please note that $p_{m_i} \in T_j$ and should satisfy the Eq. (\ref{eqn:pmin}). Overall, the set $\Upsilon$ contains all Support and Priority voxels for all tentacles and can be defined as $\Upsilon = \{S^{v}_j \cup P^{v}_j \quad | \quad j = 1,...,N^t\}$. The Fig. \ref{fig:support_priority} shows the visualization of extracted Priority and Support voxels. ~\\
\begin{subequations}
    \begin{align}
        v_i &\in
            \begin{cases}
                \label{eqn:sp_voxel}
                P^{v} & if \quad |A_p(o_i) - p_{m_i}| \leq \tau^{P^{v}} \\
                S^{v} & if \quad \tau^{P^{v}} < |A_p(o_i) - p_{m_i}| \leq \tau^{S^{v}} 
            \end{cases}
    \end{align}
    
    \begin{align}
        \label{eqn:sp_set}
        &S^{v} \cap P^{v} = \emptyset \quad \& \quad S^{v} \cup P^{v} \subseteq G
    \end{align}
    
    \begin{equation}
        \label{eqn:pmin}
        \begin{aligned}
            &|A_p(o_i) - p_{m_i}| \leq |A_p(o_i) - p_{k_j}|\text{,} \quad \forall k_j\\ &\text{where} \quad k_j = 1,...,n^s_j \quad  \text{and} \quad p_{k_j} \in T_j.
        \end{aligned}
    \end{equation}
    \label{eqn:sp_full}
\end{subequations} ~\\
 ~\\
The occupancy weight for each voxel $\beta_i$ is calculated by the function in Eq. (\ref{eqn:weight}). For $\forall v_i \in P^{v}$ the equation gives the maximum weight $\beta_{max}$, since Priority voxels are the closest ones to the corresponding trajectory and any occupancy on them might imply a high-impact collision risk. When $v_i \in S^{v}$, the value of the weight become decreasing for more distant voxels, where the rate can be adjusted by the parameter $\alpha_{\beta} > 0$.
\begin{equation} 
\begin{aligned}
\beta_i &= 
    \begin{cases} 
        \beta_{max} & if \quad v_i \in P^{v}\\
        \frac{\beta_{max}}{\alpha_{\beta} |A_p(o_i) - p_{m_i}|} & if \quad v_i \in S^{v}.
        \label{eqn:weight}
    \end{cases}
\end{aligned}
\end{equation}~\\

\subsection{Heuristic Trajectory Evaluation} \label{sec:traj_eval}
At each algorithm cycle, all trajectories are evaluated by five heuristic metrics derived from the path planning literature. We address these metrics as Navigability $\Pi^{nav}_j$, Clearance $\Pi^{clear}_j$, Nearby Clutter $\Pi^{clut}_j$, Goal Closeness $\Pi^{close}_j$ and Smoothness $\Pi^{smo}_j$. Our interpretations of these heuristic functions are given in the following subsections. ~\\

\subsubsection{Navigability} \label{sec:navigability}
For each trajectory $j$, $\Pi^{nav}_j$ assigns whether it is navigable ($1$), non-navigable ($0$) or temporarily navigable ($-1$) using the Eq. (\ref{eqn:navigability}).
\begin{equation} 
    \begin{split}
        \label{eqn:navigability}
        \Pi^{nav}_j &= 
        \begin{cases} 
            1, \hspace{0.1em} & if \quad l^{obs}_j = l^t_j\\
            0, \hspace{0.1em} & if \quad l^{obs}_j < \tau^{crash} \\
            -1, \hspace{0.1em} & if \quad \tau^{crash} < l^{obs}_j < l^t_j.
	    \end{cases}
	\end{split}
\end{equation}
Here, the crash distance threshold $\tau^{crash}$ can be adjusted as shown in Eq. (\ref{eqn:tau_crash}), by the rate parameter $\alpha_{crash}$ such that $0 < \alpha_{crash} \leq 1$. ~\\
\begin{equation} 
    \begin{split}
        \label{eqn:tau_crash}
        \tau^{crash} = \alpha^{crash} l^t_j.
	\end{split}
\end{equation}
The variable $l^{obs}_j$, which is calculated as in Eq. (\ref{eqn:lobs}), is the distance to the first occupied navigation point at $s^{obs}$ along the trajectory. In order to determine, the occupancy around a navigation point, all Priority voxels around that point are queried by the local map. Based on these queries, the occupancy counter $H_{k_j}$ is computed by starting from the closest navigation point to the robot. For each occupied Priority voxel $v_{i_j}$, the counter $H_{k_j}$ is incremented as in Eq. (\ref{eqn:counter}). If at any navigation point the value of $H_{k_j}$ pass the occupancy error threshold $\tau^{D_{err}}$, then the $k_{obs}$ is equal to the index of that point as given in Eq. (\ref{eqn:kobs}).
\begin{subequations} 
    \begin{align}
        \label{eqn:lobs}
        l^{obs}_j &= \frac{l^t_j k^{obs}_j}{n^s_j}\\ 
        \label{eqn:kobs}
        k^{obs}_j &= \min_j k_j\text{,} \quad \text{s.t.} \quad H_{k_j} > \tau^{D_{err}} \quad \forall k_j\\
	    \label{eqn:counter}
        H_{k_j} &= \sum_{v_{i_j}}^{} 1 \quad \text{s.t.} \quad v_{i_j} = (o_{i_j}, \beta_{i_j}, m_{i_j}=k_j, c_{i_j}) \in P^{v} \quad \text{and} \quad A_{\rho}(o_{i_j}) > 0
	\end{align}
\end{subequations}
\begin{figure}[!ht]
\centering
\includegraphics[width=4.5in]{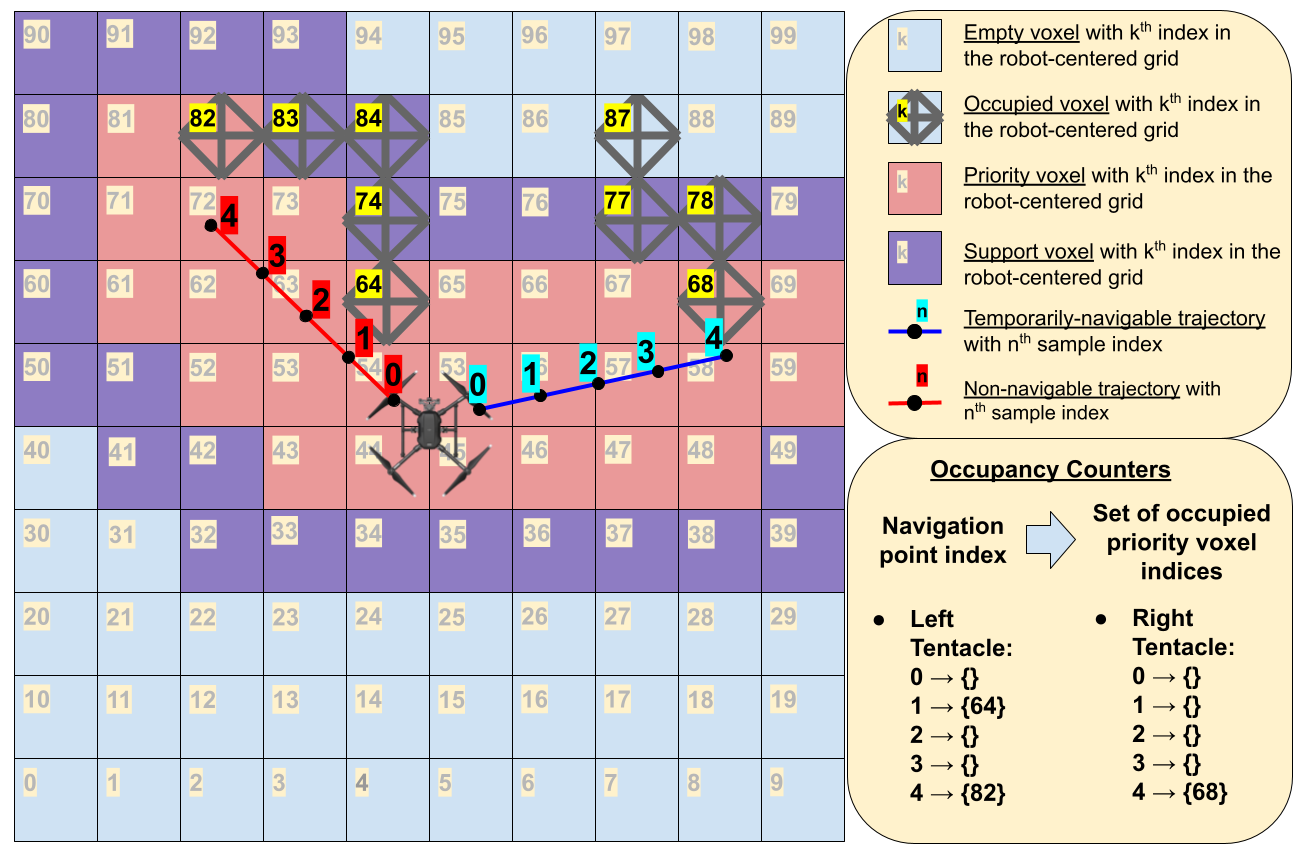}
\caption{Given a planar navigation scenario where the Support (magenta) and Priority (red) voxels are extracted for the two trajectory examples. Suppose that the crash distance is up to the second navigation point, the left trajectory becomes non-navigable since index of the navigation point (whose occupancy counter is greater than ${\tau^{D_{err}}=0}$) is less than the crash distance. On the other hand, the right trajectory is classified as temporarily navigable since the distance to the first occupied navigation point is greater than the crash distance.}
\label{fig:navigability}
\end{figure}~\\

\subsubsection{Clearance} \label{sec:clearance}
$\Pi^{clear}_j$ reflects proximity of an obstacle along the trajectory. It is obtained by the ratio of the distance to the first occupied navigation point $l^{obs}_j$, and the tentacle length $l^t_j$ as shown in Eq. (\ref{eqn:clear}). The value range of the function changes from $0$ (totally clear path) to $1$ (occupied) based on the variable $l^{obs}_j$ which is already calculated in Eq. (\ref{eqn:lobs}) while obtaining the Navigability value.
\begin{equation} 
    \label{eqn:clear}
	\begin{aligned}
	    \Pi^{clear}_j = 1 - \frac{l^{obs}_j}{l^t_j}.
	\end{aligned}
\end{equation} ~\\

\subsubsection{Nearby Clutter} \label{sec:clutter}
The nearby clutter value $\Pi^{clut}_j$ is calculated for each trajectory $j$ as in Eq. (\ref{eqn:clutter}). Here, the first variable $\Omega^{tot}_j$ equals to the sum of all Priority and Support voxel weights $\beta_i$. The second variable $\Omega^{obs}$, is defined as the weighted sum these voxels' occupancy values $A_{\rho}(o_i)$.
\begin{subequations} 
	\begin{align}
	    \Pi^{clut}_j &= \frac{\Omega^{obs}}{\Omega^{tot}}\text{,} \quad \text{where,} \\
	    \Omega^{tot} &= \sum_{v_i}^{} \beta_i\\
	    \Omega^{obs} &= \sum_{v_i}^{} \beta_i A_{\rho}(o_i)\\
	    v_i &= (o_i, \beta_i, m_i, c_i) \in P^{v} \cup S^{v}.
    \end{align}
    \label{eqn:clutter}
\end{subequations} ~\\

\subsubsection{Goal Closeness} \label{sec:closeness} $\Pi^{close}_j$ is calculated by the Euclidean distance between a specified point $p_s$ on the trajectory $i$ and the goal point $p^{goal}$ as shown in Eq. (\ref{eqn:close}). $p_s$ is selected at the crash distance $k^{obs}$ if the distance to the goal is greater than the max length of the trajectory ${l^t_j}$. If the goal distance is closer, the distance to the goal is projected on the trajectory and the nearest navigation point is selected as $p_s$. To enable objective weighting between heuristic functions, $\Pi^{close}_j$ is normalized by the maximum of all trajectories.
\begin{equation} 
	\begin{aligned}
	\label{eqn:close}
	    &\Pi^{close}_j = \frac{|^{W}p_s - ^{W}p^{goal}|}{\displaystyle\max_j \Pi^{close}_j},\\
	        &\text{where,} \quad
            p_s = 
                \begin{cases} 
                    p^{crash}, \hspace{0.1em} \quad  if \quad |^R{p^{goal}}| > l^t_j\\
                    p^{proj}, \hspace{0.1em} \quad if \quad |^R{p^{goal}}| \leq l^t_j.
	            \end{cases}
	\end{aligned}
\end{equation} ~\\
% TODO: Define p_{crash} and p_{proj}.
	
\subsubsection{Smoothness} \label{sec:smoothess}
$\Pi^{smo}_j$ function is defined to achieve smoother trajectory transitions. It assigns lower values to trajectories whose first navigation point $p_{0_j}$ is closer to the first point on the previously selected trajectory $p_{0_{prebest}}$ as shown in the Eq. (\ref{eqn:smooth}). Similar to $\Pi^{close}_j$, Smoothness is also normalized by the maximum value of all trajectories. ~\\
\begin{equation} 
	\begin{aligned}
	    \Pi^{smo}_j = \frac{|^{R}p_{0_j} - ^{R}p_{0_{prebest}}|}{\displaystyle\max_j \Pi^{smo}_j}
	    \quad \text{s.t.} \quad ^{R}p_{0_j} \in T_j, \quad ^{R}p_{0_{prebest}} \in T_{prebest}.
	\end{aligned}
	\label{eqn:smooth}
\end{equation} ~\\

\subsection{Trajectory Selection} \label{sec:traj_sel}
The cost function of each trajectory $F_j$, is calculated by the weighted sum of the four heuristic functions, $\Pi^{clear}_j, \Pi^{clut}_j, \Pi^{close}_j,\Pi^{smo}_j$ as in the Eq. (\ref{eqn:value_func}) where their weights are given as $\lambda^{clear}, \lambda^{clut}, \lambda^{close}, \lambda^{smo}$ respectively. The trajectory which is evaluated as the minimum of $F_j$ and classified as completely or temporary navigable by $\Pi^{nav}_j$ is selected as the best trajectory as in the Eq. (\ref{eqn:best_traj}).
\begin{equation}
    \label{eqn:value_func}
	\begin{aligned}
	    F_j = \lambda^{clear} \Pi^{clear}_j + \lambda^{clut} \Pi^{clut}_j \\
	+ \lambda^{close} \Pi^{close}_j + \lambda^{smo} \Pi^{smo}_j
	\end{aligned}
\end{equation}

\begin{equation}
    \label{eqn:best_traj}
	\begin{aligned}
	    j^{best} = arg\min_j F_j, \quad \forall j \quad \text{where} \quad \Pi^{nav}_j \ne 0.
	\end{aligned}
\end{equation} ~\\

\subsection{Next Pose Calculation} \label{sec:next_pose}

Our navigation framework is designed as a higher level controller which outputs a desired robot pose for the next time step. To execute the actual motion, this desired target needs to be processed by a pose controller which sends force/torque commands to the actuators of the robot. In our previous paper \cite{akmandor20203d}, we already consider the kinematic constraints of the robot, such as maximum lateral and angular speeds, to determine the next feasible robot pose. Instead of directly sending the first navigation point on the selected trajectory to the motion controller, we interpolate the point between the current robot position $^Rp=[0,0,0]$ and the navigation point at the crash distance $^Rp_{{k^{obs}}_{j^{best}}}$. At each iteration, we send these desired pose commands to a lower-level controller for the motion execution. ~\\
\\
As we discuss in Sec. (\ref{sec:sample_traj}), navigation points on selected trajectories may not be feasible for the dynamical system to reach within the the algorithm's processing time, $d_t$. Even if it is feasible, the parameters of the controller might be tuned such that the desired pose cannot be executable in $d_t$. To reduce the feasibility gap between the low level controller and our algorithm, we add two nominal speed variables, $\alpha^{\omega}$ and $\mu^{nom}$, to achieve the desired robot pose in the next time step. The given angular velocity weight $\alpha^{\omega}$ regulates the orientation while the nominal lateral velocity $\mu^{nom}$ adjusts the rate of the controller reaching to the desired next position. $\mu^{nom}$ is automatically adjusted based on the conditions as shown in Algorithm \ref{algo_nextpose}.
 ~\\
 \begin{algorithm}[ht!]
    \SetKwData{Begin}{Begin}
    \SetKwFunction{Begin}{Begin}
    \SetKwInOut{Input}{Input}
    \SetKwInOut{Output}{Output}
    \SetKwProg{Main}{main()}{}{end}
    \SetKwProg{Begin}{begin}{}{end}
    \SetKwProg{Ifend}{if}{}{end}
    \SetKwProg{If}{if}{}{}
    \SetKwProg{Else}{else}{}{end}
    \SetKwProg{End}{}{}{end}
    \SetKwProg{Elseif}{elseif}{}{end}
    \SetKwFunction{FRecurs}{FnRecursive}%

    \Input{Robot parameters $\chi^{R}$, \\
    Index of the best trajectory $j^{best}$, \\
    Angular velocity weight $\alpha^{\omega}$ \\
    Current lateral velocity $\mu_t$, \\
    Nominal lateral velocity $\mu^{nom}$, \\
    Rate of the lateral velocity $\Delta\mu$, \\
    }
    \Output{Next pose of the robot $[p^{next}, q^{next}]$}
    \Begin{}
    {
        %$\varphi^{nom} \leftarrow \omega^{nom}_{\varphi} d_t$ \\
        %$d_{\varphi} \leftarrow \frac{\varphi^{max} - \varphi^{min}}{n_{\varphi}}$ \\
        %$\kappa_\varphi \leftarrow n_{\varphi} - (n_{\varphi}-1) \frac{|\varphi^{next}|}{0.5\varphi}$ \\
        //// Calculate the desired robot orientation for the next time step. \\
        $\varphi^{max} \leftarrow \omega^{max}_{\varphi} d_t$ \\
        $\varphi^{next} \leftarrow \arctan(y^{best}, x^{best})$ where $p_0{_j^{best}}=[x^{best}, y^{best}, z^{best}]$ \\
        \Ifend{ $|\varphi^{next}| > \varphi^{max}$}
        {
            $\varphi^{next} = \varphi^{max} \frac{\varphi^{next}}{|\varphi^{next}|}$
        }
        $\varphi^{next} \leftarrow \alpha^{\omega} \varphi^{next}$ \\
        $q^{next} \leftarrow eulerToQuaternion([0, 0, \varphi^{next}])$ \\
        //// Calculate the desired robot position for the next time step. \\
        // Condition 1: \\
        \If{$\mu^{nom} \geq \mu_t$}
        {
            \If{$\mu^{nom} - \mu_t > \Delta\mu$}
            {
                $\mu_t \leftarrow \mu_t + \Delta\mu$ \\
            }
            \Else{}
            {
                $\mu_t \leftarrow \mu^{nom}$ \\
            }            
        }
        \Else{}
        {
            \If{$\mu_t - \mu^{nom} > \Delta\mu$}
            {
                $\mu_t \leftarrow \mu_t - \Delta\mu$ \\
            }
            \Else{}
            {
                $\mu_t \leftarrow \mu^{nom}$ \\
            }
        }
        // Condition 2: \\
        \Ifend{$\|p^{goal}\| < 0.25 l^t$}
        {
            $\mu_t \leftarrow \mu_t - 2\Delta\mu$ \\
        }
        // Condition 3: \\
        \If{$\mu_t > \mu^{max}$}
        {
            $\mu_t \leftarrow \mu^{max}$
        }
        \Elseif{$\mu_t < \mu^{min}$}
        {
            $\mu_t \leftarrow \mu^{min}$
        }
        $\chi^{nom} \leftarrow \mu^{nom} d_t$ \\
        $p^{next} \leftarrow linearInterpolation([0,0,0], ^Rp_{{k^{obs}}_{j^{best}}}, \chi^{nom})$
    }
    \caption{CalculateNextPose}
    \label{algo_nextpose}
\end{algorithm}

\subsection{Implementation Details} \label{sec:implement}
We implement the proposed framework and the required data structures in ROS environment using C++. The pseudo-code is demonstrated in Algorithm \ref{algo} which enables a mobile robot to navigate in an unknown environment given the input goal pose(s). We assume that the global positioning and odometry information of the robot are available throughout the navigation. To avoid obstacles on the map, occupancy data needs to be available at a frequency which is equal or higher than the main loop's rate. ~\\
\begin{algorithm}[ht!]
    \SetKwData{Left}{left}
    \SetKwData{Begin}{Begin}
    \SetKwData{Up}{up}
    \SetKwFunction{Union}{Union}
    \SetKwFunction{Begin}{Begin}
    \SetKwInOut{Input}{Input}
    \SetKwInOut{Output}{Output}
    \SetKwProg{Main}{main()}{}{end}
    \SetKwProg{Begin}{begin}{}{end}
    \SetKwFunction{FRecurs}{FnRecursive}%

    \Input{Array of goal poses $A_g$,\\
    Point cloud data $D$,\\
    Robot parameters $\chi^{R}$,\\
    Offline parameters $\chi^{off}$,\\
    Online parameters $\chi^{on}$}
    \Output{Target pose of the robot $[p^{next}, q^{next}]$}
    \Begin{}
    {
        $A_p, A_{\rho} \leftarrow$ InitializePositionOccupancyArrays($\chi^{R}$, $\chi^{off}$); \\
        $Q \leftarrow$ GenerateSampledTrajectories($\chi^{R}$, $\chi^{off}$); \\
    	$\Upsilon \leftarrow$ ExtractSupportPriorityVoxels($\chi^{off}$, $Q$); \\
    	\While{$goalNotReached(A_g)$ and $timeLimitNotReached$ and $notFailed$}
    	{
    	    $M \leftarrow$ UpdateLocalMap($D$); \\
    		\For{each trajectory $j$}
    		{
    		    $H_j, \Omega^{tot}_j, \Omega^{obs}_j \leftarrow$ UpdateOccInfo($\chi^{off}$, $\chi^{on}$, $\Upsilon$, $A_p$, $A_{\rho}$); \\
    		    $\Pi^{nav}_j \leftarrow$ UpdateNavigability($\chi^{off}$, $\chi^{on}$, $H_j$); \\
    			$\Pi^{clear}_j \leftarrow$ UpdateClearance($\chi^{off}$, $\Pi^{nav}_j$); \\
    			$\Pi^{clut}_j \leftarrow$ UpdateClutter($\Omega^{tot}_j$, $\Omega^{obs}_j$); \\
    			$\Pi^{close}_j \leftarrow$ UpdateCloseness($\chi^{R}$, $\chi^{on}$, $A_g$); \\
    			$\Pi^{smo}_j \leftarrow$ UpdateSmoothness($\chi^{R}$, $j^{best}$); \\

    			$F_{j} \leftarrow$ UpdateCost($\Pi^{nav}_j$, $\Pi^{clear}_j$, $\Pi^{clut}_j$, $\Pi^{close}_j$, $\Pi^{smo}_j$); \\
    	        $j^{best} \leftarrow$ SelectBestTrajectory($F_j$); \\
    	    }
    		$P^{R} \leftarrow$ CalculateNextPose($\chi^{R}$, $j^{best}$, $\alpha^{\omega}$, $\mu_t$, $\mu^{nom}$, $\Delta\mu$); \\
    	}
    }
    \caption{Reactive navigation by heuristically evaluated pre-sampled trajectories}
    \label{algo}
\end{algorithm}

\subsubsection{Parameters} \label{sec:parameters}
Structure of robot parameters $\chi^R$ includes volumetric, and kinematic information of the robot along with occupancy sensor specifications as described in Table \ref{table:params}. In order to enable utilization across robotic platforms, instead of considering exact volume of the robot, we adopt a bounding box model. The resolution and the range information of the navigation sensor define the size of the robot-centered 3D grid. The maximum lateral and angular velocity parameters used to calculate the desired next pose of the robot. ~\\
\begin{table}[!ht]
	\caption{Parameters}
	\begin{center}\label{table:params}
	    \begin{tabular}{|p{4cm} | p{8cm}|}
	        \hline
			{\bf Robot Parameters $\chi^{R}$} & {\bf Description}\\[0.5pt]
			\hline
            ${w^R, {l^R}, {h^R}}$ & Width, length, height of the robot \\
            \hline
            ${\mu^{max}}$ & Max forward lateral velocity of the robot \\
            \hline
            ${\omega_{\varphi}, \omega_{\theta}, \omega_{\psi}}$ & Max angular velocity of the robot in yaw, pitch and roll \\
            \hline
            $d^s$ & Resolution of the navigation sensor \\
            \hline
            $\rho_x, \rho_y, \rho_z$ & Max range of the navigation sensor in $x$, $y$ and $z$ axes \\
			\hline
			\hline
			{\bf Offline Parameters $\chi^{off}$} & \\[0.5pt]
			\hline
			${d^v}$ & Voxel dimension\\
            \hline
            $n^{v}_{\{x,y,z\}}$ & Number of grid voxels in each axes\\
            \hline
            $n_{\varphi}, n_{\theta}, n_{\psi}$ & Number of trajectories in yaw-pitch-roll\\
            \hline
            $n^s$ & Number of sampling points on a tentacle\\
            \hline
            $l^t_{max}$ & Max trajectory length\\
            \hline
            $\varphi, \theta, \psi$ & Covered angle of trajectories in yaw, pitch and roll\\
            \hline
            ${\tau^{P}}, {\tau^{S}}$ & Distance thresholds of Priority and Support voxels\\
			\hline
			$\beta_{max}$ & Max occupancy weight of Priority voxels\\
			\hline
			$\alpha_{\beta}$ & Occupancy weight scale of Priority and Support voxels\\
			\hline
			\hline
			{\bf Online Parameters $\chi^{on}$} & \\[0.5pt]
			\hline
			$\alpha^{crash}$ & Crash distance scale \\
            \hline
            $\lambda^{clear}$ & Clearance weight \\
            \hline
            $\lambda^{clut}$ & Nearby clutter weight \\
            \hline
            $\lambda^{close}$ & Goal closeness weight\\
            \hline
            $\lambda^{smo}$ & Smoothness weight\\
            \hline
            $\alpha^{\omega}$ & Angular velocity weight\\
            \hline
            $\mu^{nom}$ & Nominal lateral velocity\\
            \hline
            $\Delta\mu$ & Rate of the lateral velocity\\
            \hline
			\end{tabular}
	\end{center}
\end{table} ~\\
The remaining input parameters, which directly affect the performance of the proposed navigation algorithm, are grouped into two categories and named as offline $\chi^{off}$ and online $\chi^{on}$ as given in Table \ref{table:params}. Since the reactive nature of the algorithm is empowered by the fast computation, the offline parameters are adjusted only before the navigation. On the other hand, online parameters can be updated during the navigation without causing much computational burden but to improve the performance. In essence, the general form of the tentacles and the robot-centered grid are formed by $\chi^{off}$ while navigation preferences such as greediness towards the goal or timidness while avoiding obstacles are tuned by $\chi^{on}$.~\\

\subsubsection{Main Algorithm} \label{sec:main_algo}
The algorithm begins with the initialization of the robot-centered grid structure, which consist of a two linear array of size $(N^v)$. First array $A_p$ keeps positions of the voxel centers to enable mapping between 3D coordinates to linear indices. The second one $A_{\rho}$ allocates memory for the occupancy information. The geometric structure of trajectories are generated by defining navigation points considering the dimensions of the 3D grid. All navigation points are stored in a 2D vector of size $N^t$ by $n^s_j$, where each row of points are sampled from the same trajectory. Having the robot-centered grid and navigation points, Support and Priority voxels $v=(o, \beta, m, c)$ are extracted and kept in a 2D array of size $N^t$ by $n^{SP}_j$ where $n^{SP}_j \subseteq N^v$. Hence, the order of growth of the initialization process can be given as $O(N^t n^{v})$. ~\\
\\
The main loop of the reactive navigation algorithm iterates until all goals are reached. The algorithm stops when the navigation is failed by either reaching the time limit or the robot crashes into an obstacle. Each iteration starts by updating the local map with the most recent point cloud $D$, which contains $N^D$ data point. This takes $O(N^D)$ processing time in our implementation, where we use functions from Octomap library. Having the latest local map, we computed the occupancy information around each trajectory $j$, where $j \in \{1,...,N^t\}$. This process depends on the number of extracted Priority and Support voxels and takes $O(n^{SP}_j)$ for each trajectory. Then, heuristic functions are calculated using the updated occupancy information. Therefore, computation time of all cost functions is bounded by $O(N^t \sum_{j} n^{SP}_j)$ and selecting best tentacle takes $O(N^t)$. In the last step of each iteration, targeted next robot pose is calculated as explained in \ref{sec:next_pose}. The execution of the desired pose is performed by the controller developed by \cite{lee2010geometric} to generate the rotor actuation of the UAV in the physics-based simulations. ~\\

\section{Results} \label{sec:results}

\begin{figure}[!ht]
    \centering
    \includegraphics[width=4.5in]{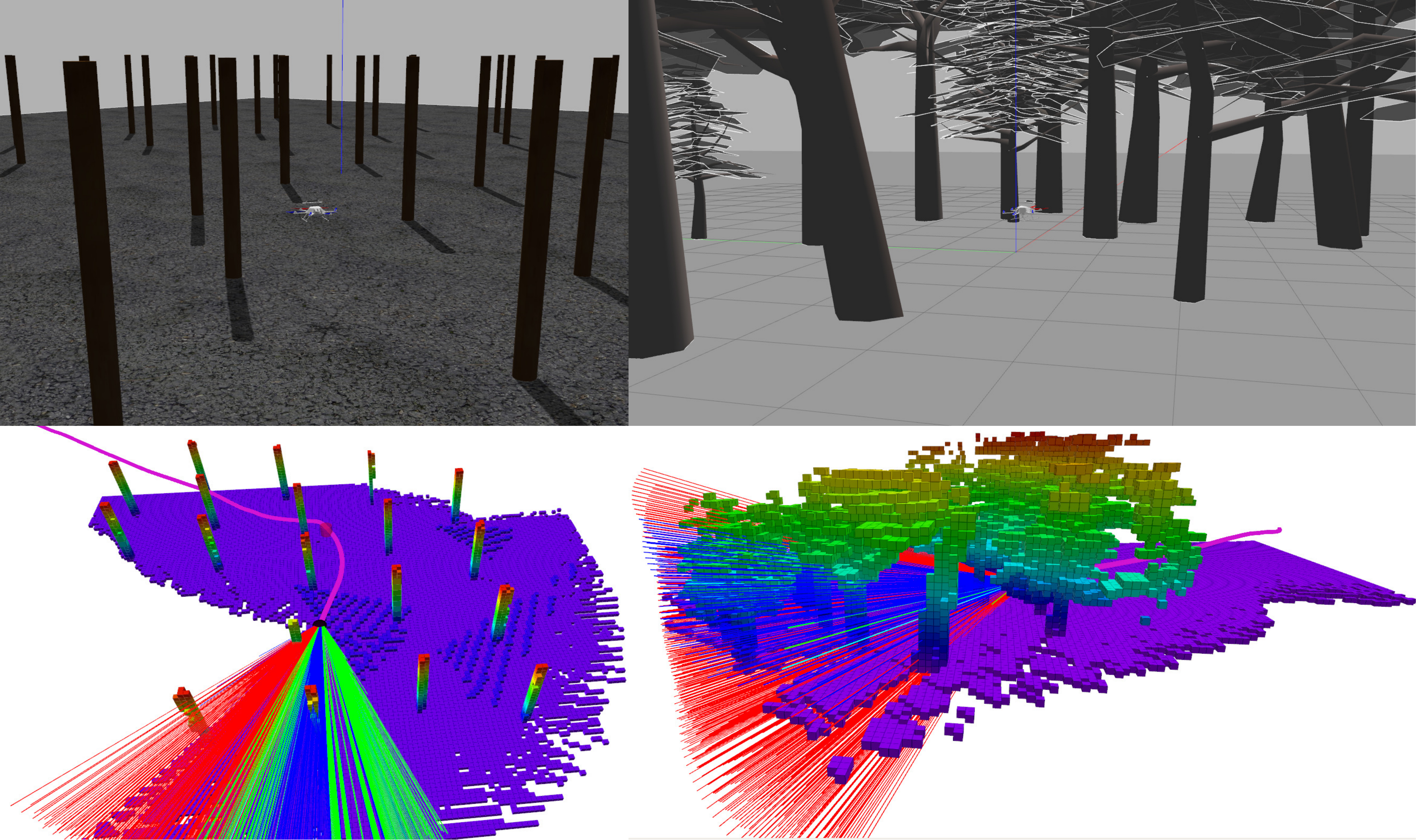}
    \caption{For the benchmark, two types of maps in Gazebo environment are used. (Top left) The first type  consists of cylindrical obstacles in $20x20m^2$ area. (Top right) The second type of map contains tree shaped obstacles whose density is $0.2trees/{m^2}$ inside of a $10x10m^2$ area. (Bottom) Rviz is used to observe status of the navigation including the local occupancy, the path of the robot, and navigability of pre-sampled trajectories.}
    \label{fig:simulation}
\end{figure}

\begin{figure}[!ht]
\centering
\includegraphics[width=4.5in]{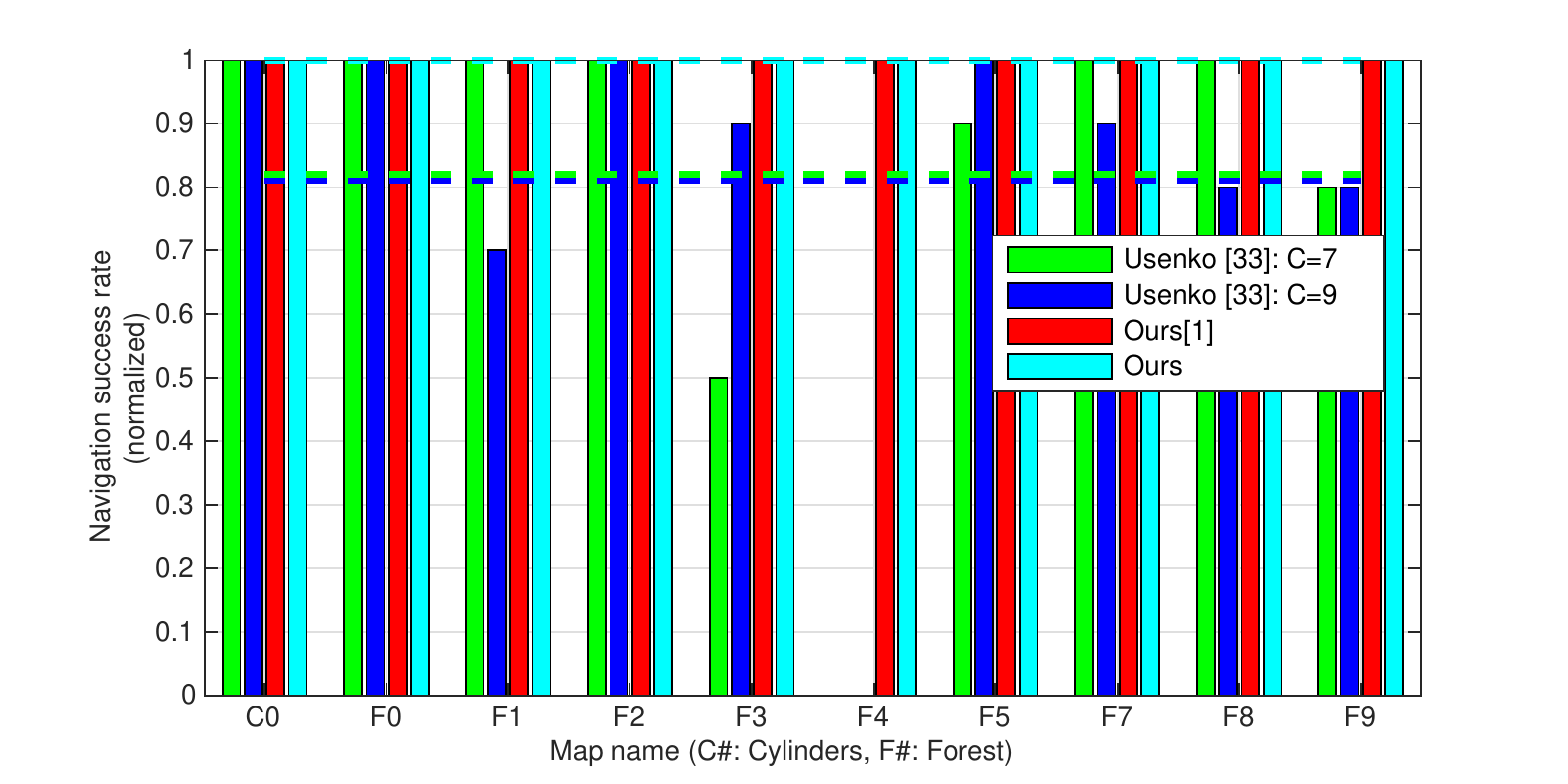}
\includegraphics[width=4.5in]{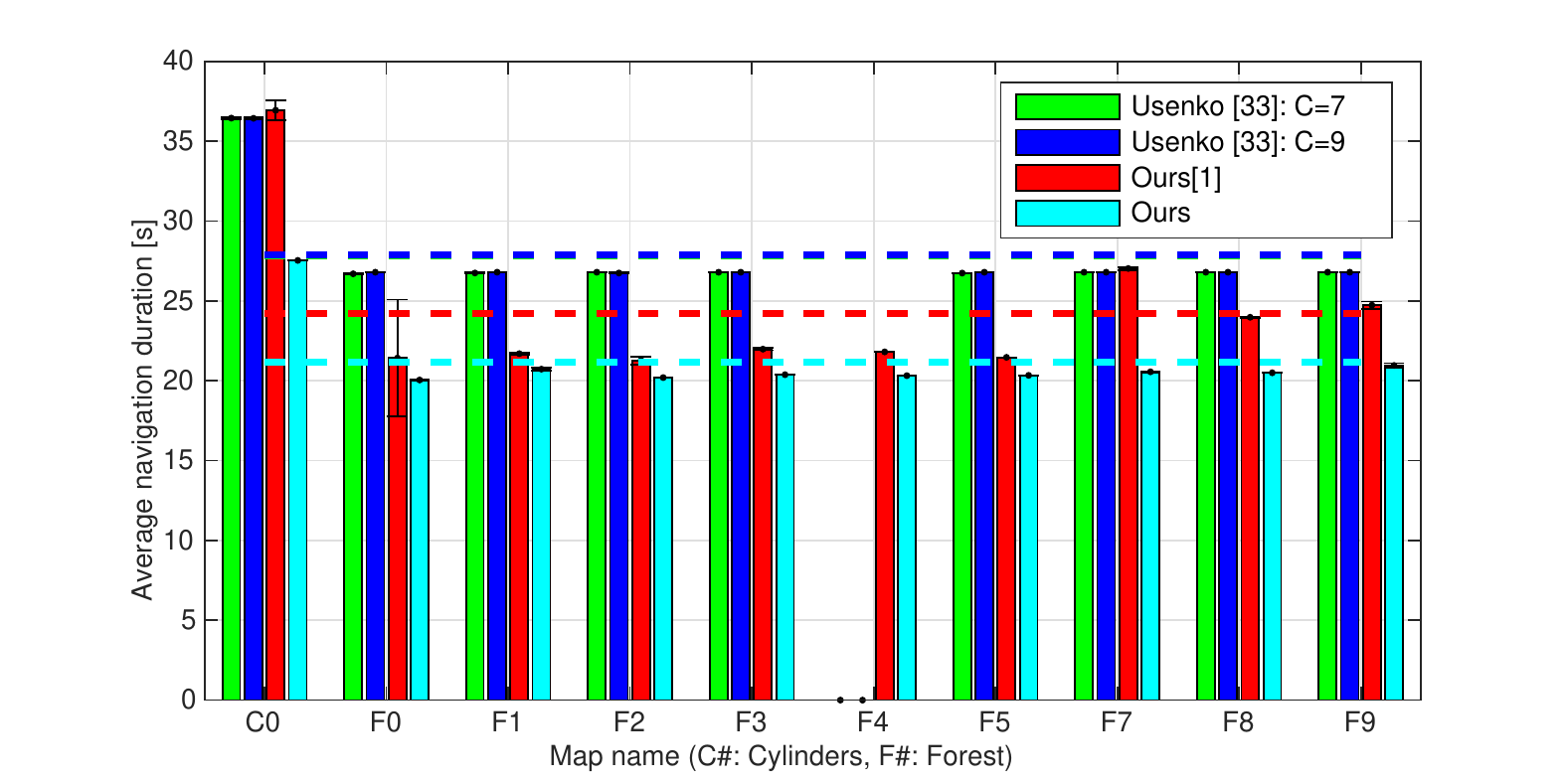}
\includegraphics[width=4.5in]{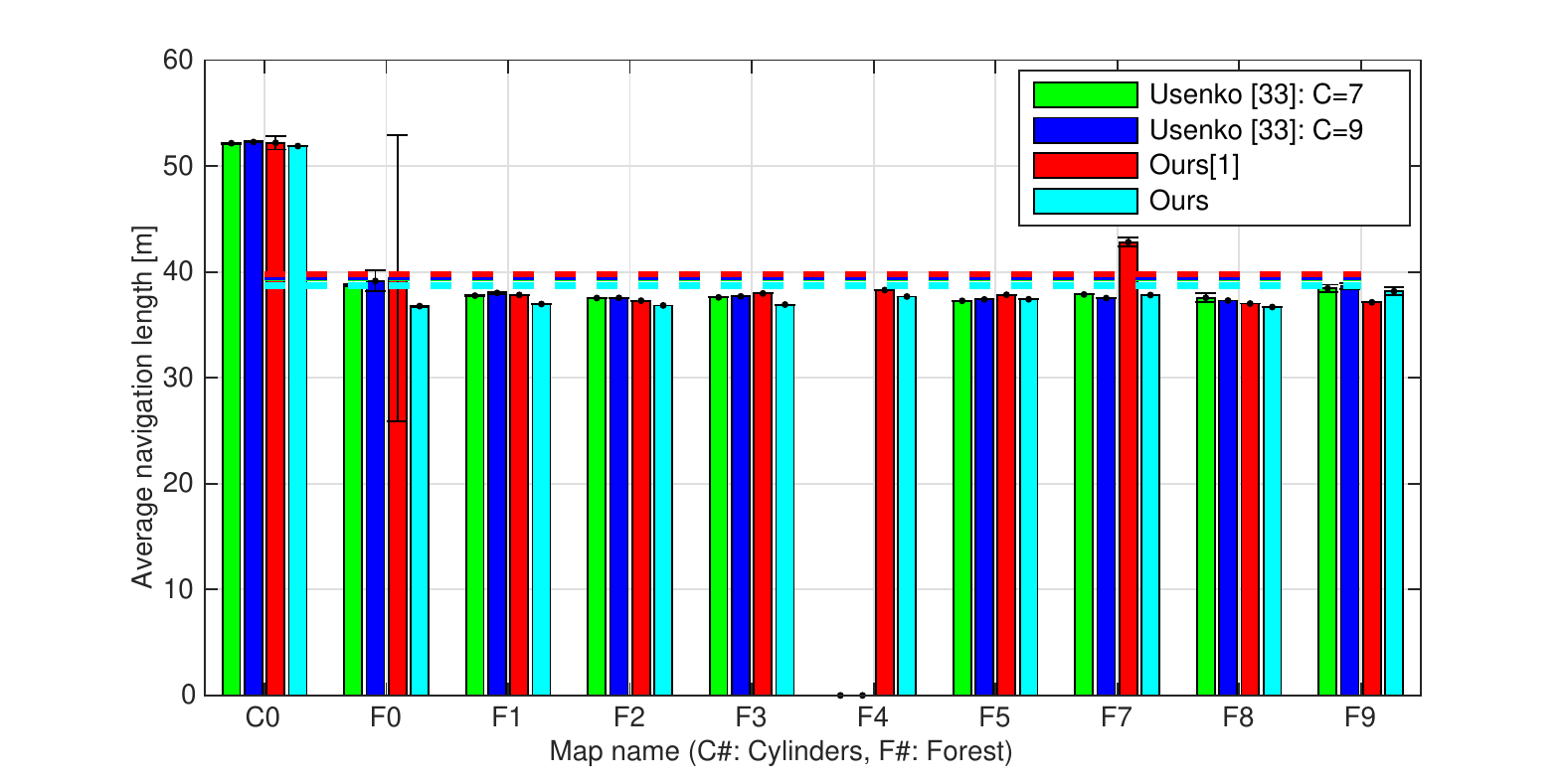}
\caption{Based on (top) navigation success rate, (middle) duration and (bottom) the path length, the statistical performances of the proposed algorithm in 10 different maps are benchmarked with the our previous implementation \cite{akmandor20203d} and the work in \cite{usenko2017real}. The dashed lines shows the average values of all maps.}
\label{fig:benchmark}
\end{figure}

 For the benchmark tests, we use two types of maps in Gazebo environment, which are available in the code repository of \cite{usenko2017real}. The first type, shown in the top left of the Fig. \ref{fig:simulation}, consists of cylindrical obstacles in $20x20m^2$ area. We keep the same goal positions, as in \cite{usenko2017real}, which are determined to maximize the travelled distance. The second type of map, provided by the ``$forest\_gen$'' ROS package \cite{oleynikova2016continuous}, contains tree shaped obstacles whose density is $0.2trees/{m^2}$ inside of a $10x10m^2$ area. We test our algorithm with physics based simulations that considers the rotor dynamics of an hexacopter. The AscTec Firefly model is used from the ``$rotors\_simulator$'' package \cite{furrer2016rotors} where the RGB-D sensor is mounted on the robot. ~\\
 ~\\
Before running the simulations, the $\chi^{R}, \chi^{on}, \chi^{off}$ are adjusted considering the robot's kinematics, sensor specifications and the navigation strategy (such as greediness to reach the goal vs. timidness while avoiding obstacles). The range of the sensor regulates the maximum trajectory length and covered angles along yaw and pitch. Hence, $l^t_{max}, \varphi, \theta$ are set to $10m$, $60^o$ and $45^o$ respectively. The priority distance threshold $\tau^{P}$ is adjusted to $0.35m$ to encircle the bounding box of the robot while the support distance is set as $\tau^{P} = 0.5m$ empirically. Having specified the maximum tentacle length and priority distance, the number of navigation points for each trajectory is calculated by $n^s = l^t_j / \tau^{P}$ in order to keep the robot inside of the priority voxels along the trajectory. The occupancy weight scale of the priority and support voxels is adjusted to $\alpha_{\beta}=10$. The max occupancy weight is set to $\beta_{max}=1$ to keep the occupancy weights in the range of $[0,1]$. ~\\
\\
To analyze the effect of the remaining offline parameters on the computation time, three sets of simulations are performed and the results are given in the Table \ref{table:analysis}. Having the sensor with the resolution of $0.15m$, we test the voxel dimension $d^v$ for $0.2m$ and $0.1m$. In order to match the grid dimensions with the tentacles' length, the number of voxels in each axis $n^v_{x,y,z}$ are doubled when the $d^v$ is scaled down to half. This increases the total number of voxels in the grid by 8 times. Reflectively, the computation time of the initialization process of the linear arrays, when $d^v=0.1$, is measured 8 times more than when $d^v=0.2$. The second and third column of the Table \ref{table:analysis} demonstrate the linear relationship between the number of trajectories and the total computation time of the ``GenerateSampledTrajectories'' and ``ExtractSupportPriorityVoxels'' steps. As expected, the computation time is measured twice as much when the $N^t$ is doubled. The duration of the main iteration steps, especially for the ``UpdateOccInfo'' and the steps which heuristic values are updated are harder to analyze, since the calculations also depends on the momentary sensor data. The statistical computation times, shown in Table \ref{table:analysis}, indicate logical results with respect to the changes in $d^v$ and $n^v_{x,y,z}$. Moreover, the total processing time of each simulation set proves that the algorithm is capable of running within the range of frequency from $10$ to $60$ $Hz$ successfully, mostly depending on the occupancy information updates. ~\\
\begin{table}[!ht]
	\caption{Average computation time statistics of the initialization and the main iteration steps of the algorithm with respect to the voxel dimension $d^v$, the number of grid voxels in each axes $n^{v}_{\{x,y,z\}}$ and the number of trajectories $N^t$}
	\begin{center}\label{table:analysis}
	    \begin{tabular}{|p{5.5cm}|p{2cm}|p{2cm}|p{2cm}|}
	        \hline
			{\bf Initialization Steps} & {\bf Time [s]} & {\bf Time [s]} & {\bf Time [s]} \\[0.5pt]
			& {\bf $d^v=0.2$} & {\bf $d^v=0.1$} & {\bf $d^v=0.1$} \\
			& {\bf $n^v_{x,y,z}=110$} & {\bf $n^v_{x,y,z}=220$} & {\bf $n^v_{x,y,z}=220$} \\
			& {\bf $N^t=651$} & {\bf $N^t=651$} & {\bf $N^t=1271$} \\
			\hline
            InitializePositionOccupancyArrays & $0.02$ & $0.12$ & $0.16$ \\
            \hline
            GenerateSampledTrajectories + & $0.81$ & $6.03$ & $12.07$ \\
            ExtractSupportPriorityVoxels & & & \\
            \hline
            \hline
            {\bf Total} & $0.83$ & $6.15$ & $12.22$ \\
            \hline
            \hline
            {\bf Main Iteration Steps} & {\bf Time [ms]} & {\bf Time [ms]} & {\bf Time [ms]} \\[0.5pt]
            \hline
            UpdateLocalMap & $4.63$ & $9.02$ & $8.02$ \\
            \hline
            UpdateOccInfo + UpdateHeuristics & $2.81$ & $16.54$ & $31.21$ \\
            \hline
            SelectBestTentacle & $0.002$ & $0.002$ & $0.003$ \\
            \hline
            CalculateNextPose & $0.02$ & $0.02$ & $0.02$ \\
            \hline
            \hline
            {\bf Total} & $7.45$ & $25.58$ & $39.25$ \\
            \hline
			\end{tabular}
	\end{center}
\end{table} ~\\
To emphasize our contribution in the presented work, the updated algorithm is compared with the results of our previous implementation \cite{akmandor20203d} and the work in \cite{usenko2017real} within the same 10 maps (cylinders map + 9 forest maps) as shown in Fig. \ref{fig:benchmark}. Robustness of the implementation is tested by running all configurations 10 times for each map without changing any parameter. ~\\
\\
As discussed earlier in this section, most of the offline parameters can be adjusted by the given robot properties and the occupancy sensor specifications. In the benchmarks, we set the same offline parameters that are used to produce the result of the middle column in Table \ref{table:analysis} and keep them fixed for all maps. Although, we empirically tune the online parameters similar to our previous approach, this time we able to use the same online parameters for all maps, while noticeably improving the navigation performance. Clearly, the method explained in \ref{sec:next_pose} makes the algorithm more generally applicable to different navigation environments and scenarios. Yet, it is also plausible to state that manually tuning the online parameters is a weakness of the algorithm. Considering our experience on tuning these parameters, we argue that the online tuning process can be learned from the previous experiences of the robot and automatically tuned during the navigation.  ~\\
\\
Overall, the simulation results reveal that our proposed algorithm has higher success rate and enables faster navigation for all performance metrics in Fig. \ref{fig:benchmark}. Remarkably, both of our versions succeed in successive trials for all maps, while both of the configurations of the state-of-art algorithm \cite{usenko2017real} fail in the ``Forest4'' map. Noting that our algorithm is purely reactive and does not keep a global map, struggling around local minima is quite expected.

\section{Conclusion}
In this paper, we present a reactive navigation algorithm which does not rely on a global map. This is achieved by heuristic evaluations of a pre-determined group of points which sample the navigation space. The robot-centered grid structure is formed to enable fast queries of the occupancy information which is kept in a local map. In order to evaluate the trajectories and select the best possible next pose, five heuristic functions are defined. This paper also introduces a method which improves feasibility of the selected target pose at each iteration. The offline and online parameters enable adaptability of the algorithm to different environments. The approach of tuning these parameters are explained as well as the other implementation details, including computational complexity analysis. We perform physics-based simulations for the benchmark tests. Overall, the proposed algorithm outperforms two configurations of a state-of-art method and our previous version in terms of success rate, navigation length and duration.

\section*{Acknowledgment}
This research was supported by the Department of Homeland Security as part of the National Infrastructure Protection Plan (NIPP) Security and Resilience Challenge and the Northeastern University's Global Resilience Institute seed funding program. This research is also supported by the National Science Foundation under Award No. 1928654, 1935337, 1944453.

%%%% Authors are advised to submit their bibtex database files.
\bibliographystyle{splncs04}
\bibliography{main}

\end{document}